\documentclass{article}
\usepackage{spconf,amsmath,graphicx,amssymb}
\usepackage{xcolor}

\title{STRUCTURED CITATION TREND PREDICTION USING GRAPH NEURAL NETWORKS}
\name{Daniel Cummings, Marcel Nassar}
\address{Intel Corporation, USA}
\newcommand{\graph}[1]{\mathcal{#1}}
\newcommand{\set}[1]{\mathcal{#1}}
\DeclareMathOperator{\red}{\texttt{red}}
\DeclareMathOperator{\meanop}{\texttt{mean}}
\DeclareMathOperator{\sumop}{\texttt{sum}}
\DeclareMathOperator{\maxop}{\texttt{max}}
\DeclareMathOperator{\mlp}{\texttt{mlp}}
\DeclareMathOperator{\softmax}{\texttt{softmax}}

\makeatletter
\renewcommand{\paragraph}{%
  \@startsection{paragraph}{4}%
  {\z@}{0.25ex \@plus 0.25ex \@minus .5ex}{-1em}%
  {\normalfont\normalsize\bfseries}%
}
\makeatother

\begin{document}
\onecolumn
\noindent© 2020 IEEE.  Personal use of this material is permitted.  Permission from IEEE must be obtained for all other uses, in any current or future media, including reprinting/republishing this material for advertising or promotional purposes, creating new collective works, for resale or redistribution to servers or lists, or reuse of any copyrighted component of this work in other works.
\newpage
\twocolumn
%\ninept
%
\maketitle
\begin{abstract}
Academic citation graphs represent citation relationships between publications  across the full range of academic fields. Top cited papers typically reveal future trends in their corresponding domains which is of importance to both researchers and practitioners. Prior citation prediction methods often require initial citation trends to be established and do not take advantage of the recent advancements in graph neural networks (GNNs). We present GNN-based architecture that predicts the top set of papers at the time of publication. For experiments, we curate a set of academic citation graphs for a variety of conferences and show that the proposed model outperforms other classic machine learning models in terms of the F1-score.

%The modelling approach in this paper offers a solution to predict future citation impact for publications at the time of release and eliminates the need for an initial citation trend to be established. The capability offers immediate bibliometric insight to researchers across a wide range of fields. The approach could be applied other social network types since it uses feature inputs that are commonly available. 
\end{abstract}

\section{Introduction}
\label{sec:intro}
Academic citation graphs represent citation relationships between publications across the full range of academic fields and include information on authors, affiliations, references, and more. Citation graph indicators are often referred to as bibliometrics, where the author h-index, publisher ranking, and journal impact factor metrics are commonly used to gauge the importance of a publication within its field \cite{Pan14, Port17}. The interest in bibliometrics continues to grow as the number of refereed publications rapidly increases worldwide, with a recent focus predicting citation growth \cite{Egghe2005, BORNMANN2013}. Researchers benefit greatly from understanding which publications are the most relevant and influential, a.k.a. \emph{trending}, for their field of study. Influence and relevance are typically measured in terms of the number of citations or references a paper receives throughout its lifetime. Top cited papers typically reveal future trends in their corresponding domains which is of importance to both researchers and practitioners. As a result, this work focuses on predicting a set of top papers instead of focusing on the exact number of citations each paper will receive. 

Citation graphs can be represented as a graph data structure and with this in mind, we highlight the evident graph topology synergy with graph neural networks (GNNs). Recently, the popularity of GNNs has skyrocketed due to their ability to extend the impressive gains achieved by deep learning to irregular data domains such chemistry, 3D vision, relation representation learning, community detection and recommender systems \cite{Zhou18}. 

The contributions of this paper can be summarized as follows:
%The question that motivated this work was whether GNNs could be applied to citation graphs to predict the top papers at time of release. 
%To answer this question, 
(1) we curate citation graph data sets for influential conferences and (2) present a GNN-based neural network architecture which utilizes the information (authors, affiliation, citation links) of the documents that the new paper cites. (3) We validate our approach using an extensive set of experiments demonstrating the ability of this architecture to leverage the structural information present in the citation graph.  This algorithm plugs in naturally to any paper distillation and recommendation pipeline and could be applied to other social network types as well. 

\section{Prior Work}
\label{sec:prior}

It is difficult to predict top papers just from the properties of the newly published document. In prior work, most methods require citation trends to be established for up to five years after publication in order to predict what the future top papers will be \cite{ABRAM2019}. For example, a recent paper uses the journal impact factor and the citation count a year after publication as the model inputs \cite{StegehuisLW15}. A survey of research in this area show that predictors such as the numbers of authors, author impact (h-index), journal impact factor, journal past influence, citation half-life, numbers of pages, and title length are often used \cite{BAI2019407}. Other unique approaches to the problem exist that use time series forecasting, supervised link prediction, and representation learning to predict future citations between papers \cite{ABR2019,Butun2017,Asa18}. However, these works do not utilize any of the structure present in citation graphs which limits performance. 

GNNs have been applied to citations graphs such as CORA \cite{KipfW16, Velickovic2017}; however, the task there is to predict the topic of the documents using the features of the documents that it cites. Furthermore, unlike the proposed model, these models don't enforce any causality and treat all the nodes identically.

\section{Graph Neural Networks}
\label{sec:gnn}

Graph neural networks naturally lend themselves to structured prediction problems due to the neighborhoods induced by the graph topology (refer to \cite{Wu2019} for a recent survey).
A graph neural network is a parametric deep learning model that operates on graph structured data. This data is composed of a graph $\graph{G}$ and a set of node and/or edge features. 
A graph $\graph{G} \in \mathbb{G}$ is a tuple $(\set{V}, \set{E})$ denoted by $\graph{G}(\set{V}, \set{E})$ consisting of a vertex set $\set{V}=\{v_i\}_{i=1}^{N_\set{V}}$ and an edge set $\set{E}=\{e_j\}_{j=1}^{N_\set{E}}$. 
%In weighted directed graphs, each edge $e_j$ is in turn a 3-tuple $(v, u, r)$ where $v$ is the source node, $u$ is the destination node, and $r$ is the edge label, whereas each edge in an undirected graph can be represented as a 2-tuple $\left(\{v, u\}, r\right)$.
Denoting the feature of node $v_i$ by $f_i\in\mathbb{R}^N$,
a graph convolution operator $g: \mathbb{G}\times \mathbb{R}^{|\set{V}|\times N} \rightarrow \mathbb{G}\times \mathbb{R}^{|\set{V}|\times M}$ uses the graph structure to locally aggregate node features as follows:
\begin{equation}\label{eq:gconv}
g_{\graph{G}}(v_i) = \red\left(\{W_{i,j}f_j| v_j \in \delta_{\graph{G}}(v_i), f_j=s(v_j)\}\right)
\end{equation}
$\forall v_i \in \set{V}$ where $\red$ is a permutation-invariant reduction operation such as $\maxop$, $\meanop$, or $\sumop$. $\delta_{\graph{G}}(v_i)$ is the neighborhood of the node $v_i$ in $\graph{G}$. $W_{i,j} \in \mathbb{R}^{M\times N}$ is a feature weighting kernel transforming the graph's $N$-dimensional features to $M$-dimensional ones. 
A graph convolution layer can be built on top of the graph convolution operation, and stacking such layers constructs the graph convolution network (GCN). 
The form of the weighting kernel $W_{i,j}$ determines the flavor of the GCN model. The graph attention network (GAT) is a GCN model that uses the powerful attention mechanism \cite{Vaswani2017}  on the node features to construct the weighting kernel as $W_{i,j} = \alpha_{i,j}W$, where
$\alpha_{i,j}=\softmax \left(\mlp([Wf_i, Wf_j])\right)$ \cite{Velickovic2017}.

\section{Problem Statement}
\label{sec:prob-statement}

%The approach of this work is on predicting a set of top papers since we are looking to reveal future research trends. 
At the time of publication, we want to predict which papers will trend into the top percentiles of citation counts in the years afterward. We frame the problem as a classification task where we predict the top percentile of cited papers instead of focusing on the exact number of citations each paper will receive. 

Given a prior graph $\graph{G}_p=(\set{V}_p, \set{E}_p)$ representing the citation network of a corpus of documents and its node (document) labels $L_p$ representing whether a node (document) is trending or not, i.e. a node $v_i$ is trending if $L_{p,i}=1$, the goal is to develop a model to predict whether newly added documents will be trending based on their key features and their citation graph. 
Let us denote the new set of target nodes as $\set{V}_t$. Assume these nodes are related to $\set{V}_p$ through the edges $\set{E}_t$; then, the prediction problem can be formulated as finding
\begin{equation}
\label{eq:posterior}
p(L_{t,k}=1|\graph{G}_p,  \set{V}_t, \set{E}_t, L_p, f_p, f_t)
\end{equation}
for $\forall v_k \in \set{V}_t$ , where $L_{t,k}=1$ is the indicator function of whether node $k$ is trending, and $f_p$, $f_t$ are the prior and target node (document) features respectively. 

%The goal of this work is to develop a model that enables trend predictions on key features in recently added nodes for a given graph. We present a structured trend prediction approach using GNNs with graphs that have nodes encoded with a form of time comprehension (e.g. publication date). Although the application is on academic citation graphs in this work, the approach could generally be applied to any social network with time encoding. 
%As a result, this work focuses on predicting a set of top papers instead of focusing on exact number of citations each paper will receive.

\section{Proposed Architecture}
\label{sec:arch}

\begin{figure*}[t]
%
% \begin{minipage}[b]{1.0\linewidth}
  \centering
  \centerline{\includegraphics[width=.75\linewidth]{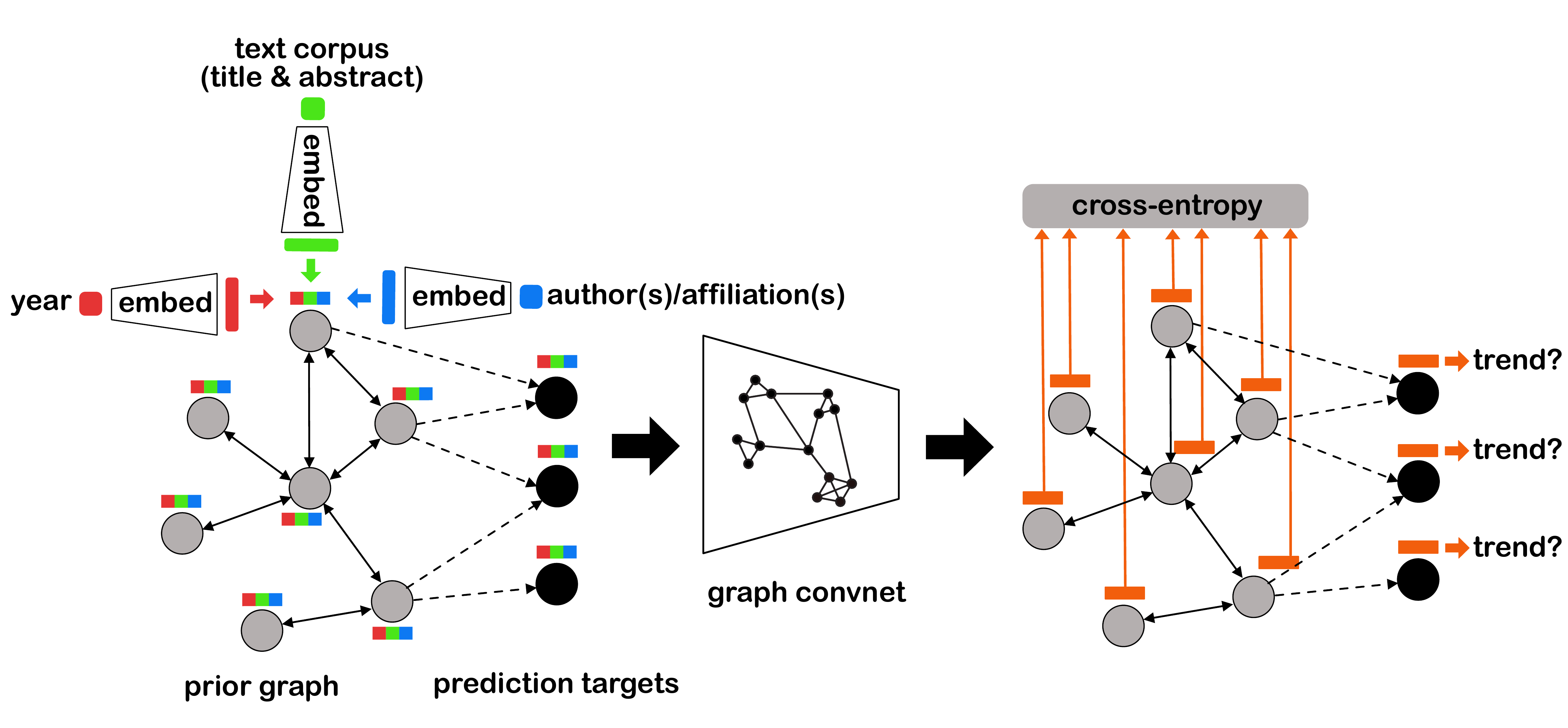}}
%  \vspace{2.0cm}
% \end{minipage}
%
\caption{Neural network architecture of the trend prediction model: gray and black nodes correspond to the prior graph nodes and prediction nodes respectively.}
\label{fig:arch}
\end{figure*}

%Based on the problem statement, our aim is to take advantage of the graph topology inherent in social networks, or specifically academic graphs in this work, by implementing a prediction model using GNNs. 
%We cast the trend prediction problem as a classification task and use the citation count as a proxy for trending property. 

Our architecture models the posterior in \eqref{eq:posterior} as
% \begin{equation}
$p(L_{t,k}=1|\graph{G}_p, \set{V}_t, \set{E}_t, L_p, f_p, f_t) = g_{\theta}(v_k, f, \graph{G})$
% \end{equation}
where $g_{\theta}(\cdot)$ is a GAT graph neural network (see Section~\ref{sec:gnn}), $\theta$ are its parameters, $\graph{G}(\set{V}_p\cup \set{V}_t, \set{E}_p\cup \set{E}_t)$ is the new union graph when the new nodes are added with their corresponding edges, and $f=[f_p, f_t]$. A key property of the proposed architecture, as seen on Fig.~\ref{fig:arch}, is that the edges are directional to impose causality constraints: the new nodes in set $\set{V}_t$ have only incident edges and thus don't contribute to the prior graph ($\graph{G}_p$) embeddings computations. As a result, the neural network computation can be decomposed into two stages: prior stage and prediction stage. The prior stage correspond to the training stage (gray nodes in Fig.~\ref{fig:arch}). The prediction phase corresponds to when the last layer of the GNN is applied to the prediction nodes (black nodes in Fig.~\ref{fig:arch}). The result of this stage can be cached and used with any arbitrary new nodes.

Feature ($f$) selection is an important choice and it is illustrated in Fig.~\ref{fig:arch}. Sparse vector representations of the text (abstract and title), author affiliation, and publication date are used as input features as they are intuitive predictors of academic topic and importance. Vectorized words are commonly used as input features in popular graph benchmark datasets \cite{Sen08}. Next, the separate sets of sparse vectors are fed into a their own embedding layers (Fig.~\ref{fig:arch}). The embedding layers are then fed into to the GCN layers which have been shown to be highly effective in graph node classification problems \cite{KipfW16}.

For a comparison baseline, a multi-layer perceptron (MLP) equivalent model was created by swapping out the GCN layers with perceptron layers. With this substitution, the MLP model loses its graph comprehension and will gives insight into how useful the graph connectivity is to the citation trend classification. Both the proposed and MLP architectures were verified to have an equal number trainable parameters.

\section{Experimental Setup}
\label{sec:experiment}
We examine the structured trend prediction model's performance on a variety of conferences that are at the cutting edge machine learning research including the IEEE International Conference on Acoustics, Speech and Signal Processing (ICASSP) conference. The source data for the various citation graphs in this work were extracted from Microsoft Academic Graph (MAG) which are also available as part of Open Academic Graph \cite{mag}. The choice of this data source was due to the large variety publication sources that were supported and the available data for author affiliations, publication date, and citation count. 

%\subsection{Citation Graph Datasets}
%\label{ssec:subhead}

The graph datasets were curated by filtering the MAG database by conference and then extracting data related to the title and abstract, author affiliations, publication date, and estimated citation count. The separate data tables were combined with tidy dataset formatting principles in mind \cite{Wick14}. For the text encoding, the title and abstract text were combined and tf-idf vectorization was applied to generate the word vectors for each paper across the graph with a maximum bound of 1000 word features. 

Table \ref{table:1} gives the node and edge count for each curated conference graph. Because the majority of citations often occur in the first three years after publication, we chose a ten year window for the prior graph $\set{G}_p$ \cite{Wang2013}. Likewise, the experiment is setup for the prediction years of 2015 and earlier since a reliable target label confidence has been established at that point. It should be noted that citation graph datasets are intrinsic to themselves and do not include every possible external citation linkage that would occur in a perfectly connected graph. 

\begin{table}[t]
\caption{Citation graph topology for the 2015 prediction year with a 10 year prior window.}
\vspace{0.2cm}
\centering
\begin{tabular}{ |p{2.6cm}|p{1.9cm}|p{1.8cm}|  }
 \hline
 
% \multicolumn{4}{|c|}{Citation graph characteristics} \\
% \hline
 Conference Graph & Nodes $(\set{V}_{all})$  & Edges $(\set{E}_{all})$ \\
 \hline
 \hline
 ICASSP     & 15813 & 20541  \\
 ICML       & 2669  & 4591   \\
 NeurIPS    & 3572  & 6345  \\
 EMNLP      & 2003  & 4563  \\
 ICIP       & 10389 & 11439 \\
 Interspeech& 5007  & 7648  \\
 ICDM       & 3750  & 4305  \\
 CIKM       & 3558  & 4629  \\
 AAAI       & 7131  & 8898  \\
 CVPR       & 7140  & 26703 \\
 \hline
\end{tabular}

\label{table:1}
\end{table}

%\subsection{Graph Formatting}
%\label{ssec:subhead}

%
%A structured time-aware approach was taken when formatting the graphs for the proposed model as shown in Figure \ref{fig:arch}. Since we want to predict the top percentiles of cited papers at the time of release, the formatted graph can be viewed as having two parts: a prior years subgraph with nodes $\set{V}_p$ and an appended prediction target year subgraph with nodes $\set{V}_t$. The prior years graph is formatted such that the edges are bidirectional across all nodes. To enforce causality, the prediction year nodes are not allowed to influence the prior subgraph; the edges are directional where the message passing only occurs from the prior to prediction year nodes. 

%\subsection{Node Target Labelling}
%\label{ssec:subhead}

The node target labels for the graphs were based on whether they fell above or below a defined citation count percentile rank. For example, if a 90\%-ile rank threshold resulted in a count of fifty, the papers with citations greater than fifty would be labelled "1" and the rest "0." Thus, we are classifying the top 10\% most-cited papers for each year in the graph in this example. The citation percentile rank labeling was performed by year since citation counts and quantiles vary significantly year-to-year. 

%As discussed in Section 5, training was performed on the prior stage and the testing/validation on the prediction stage. 
After running the model across various conference graphs, it was found that the following parameters yielded the most consistent performance: 150 epochs with learning rate of 0.001 and an L2 weight decay of 5E-4 using the Adam optimizer. For the activation function, Leaky-RELU was chosen over RELU due to the slightly faster convergence time and minor increase in accuracy. The embedding layers used 100 units for both the text and author affiliation stacks and 2 units for the publication year stack. Hence, the first GCN layer contained 202 units followed by a layer with 30 units where every layer in the model had a dropout rate of 0.1. 

\section{Results}
\label{sec:intro}
\begin{figure}[t]
\begin{minipage}[b]{1.0\linewidth}
  \centering
  \centerline{\includegraphics[width=8.5cm]{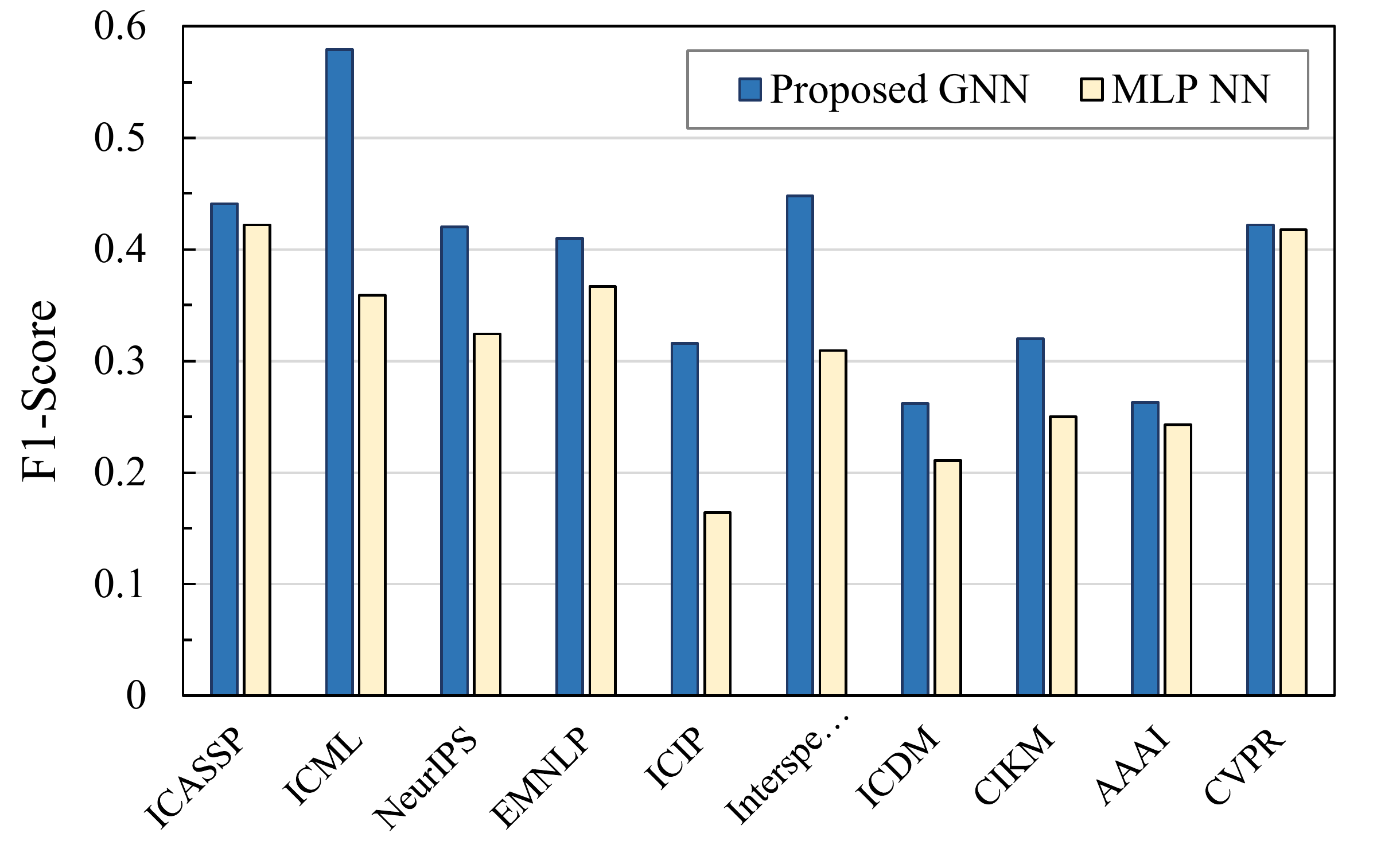}}
%  \vspace{2.0cm}
\end{minipage}
\caption{Results for predicting the top 10\% most cited papers by conference. Prediction target year 2015, 90\%-ile rank threshold.}
\label{fig:barplot}
\end{figure}

\begin{figure}[t]
\begin{minipage}[b]{1.0\linewidth}
  \centering
  \centerline{\includegraphics[width=8.5cm]{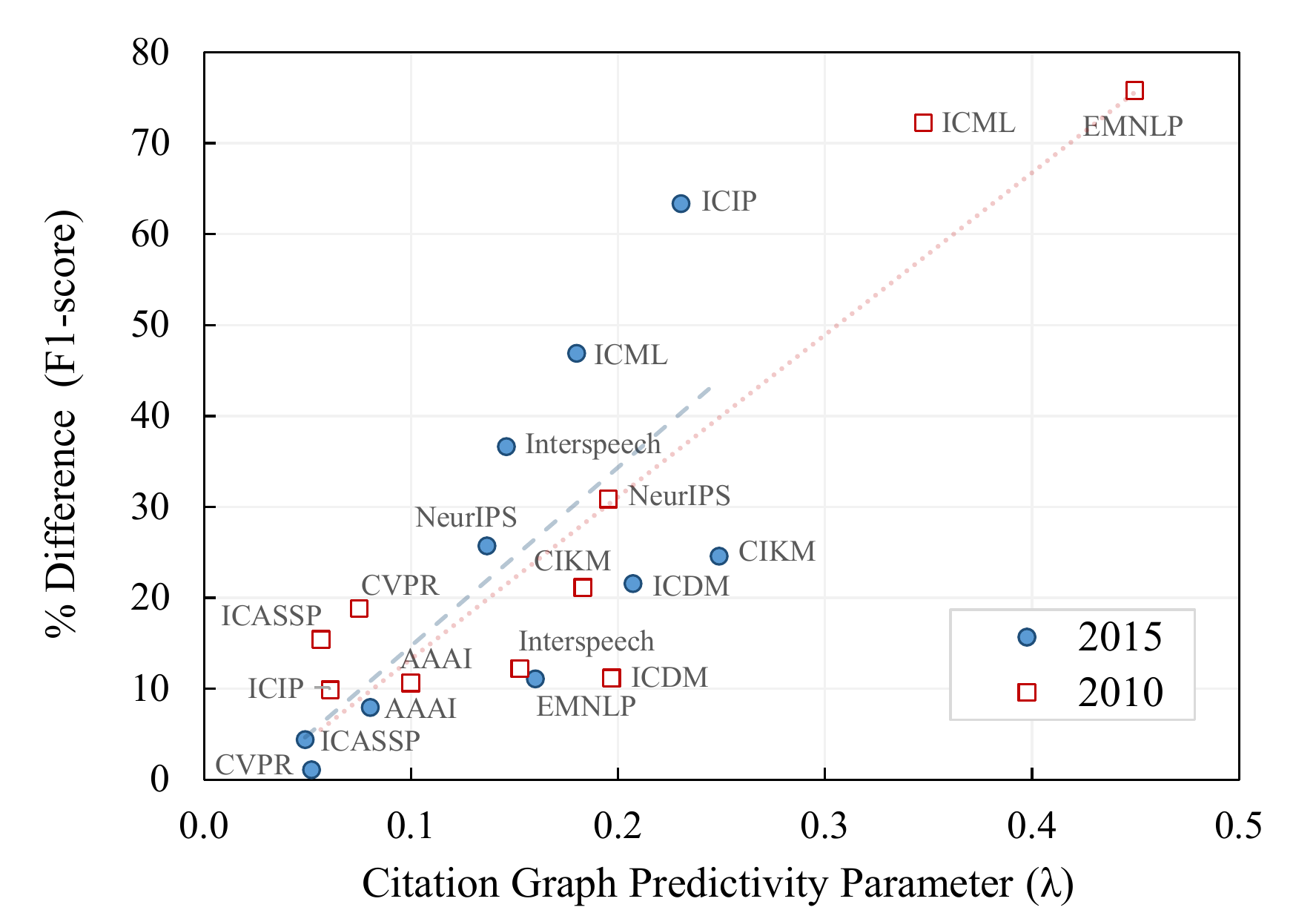}}
%  \vspace{2.0cm}
\end{minipage}
\caption{Comparison of the proposed model versus the MLP in terms of the citation graph predictivity parameter}
\label{fig:lambda}
\end{figure}

Figure \ref{fig:barplot} shows the F1-score results for the proposed trend prediction GNN model and the MLP equivalent across various conferences. The F1-score metric was chosen due to the highly unbalanced labelling of the dataset since nodes are labelled according to the citation percentile rank thresholds. The F1-score is given by:
\begin{equation}\label{eq:f1score}
F_{1} = 2\cdot \left(\frac{precision \cdot recall}{precision+recall}\right)
\end{equation}
where a perfect score is 1. The proposed model outperforms the MLP equivalent with varying levels of success. For example, ICASSP and Computer Vision and Pattern Recognition (CVPR) have only a minor benefit when using the proposed model whereas the International Conference on Machine Learning (ICML), Interspeech, and International Conference on Image Processing (ICIP) show a high F1-score benefit. 
%The precision (positive predictive value) is the ratio of true positives over the true positives plus false positives and the recall (sensitivity) is the ratio of true positives over the true positives plus false negatives.
The F1-score differences in Fig. \ref{fig:barplot} are of particular interest since they indicate how much the proposed model benefits from the use of GCN layers and connections in the graph topology. Correlations relating to the percent difference between the proposed and MLP model were searched for in terms of edge count, node count, average edges per node, and more. A moderate correlation was found when looking at the ratio of edges in the prior subgraph $\set{E}_{p}$ and the total node count $\set{V}_{all} = (\set{V}_p\cup \set{V}_t)$ as

\begin{equation}\label{eq:gcindex}
{\lambda} = \left(\frac{1}{\set{V}_{all}}\right)\left(\frac{\set{E}_{p}}{\set{E}_{all}-\set{E}_{p}}\right)\times 100
\end{equation}
where we define $\lambda$ as a citation graph predictivity parameter. The $(1/\set{V}_{all})$ term gives a lower value as the node count increases, but is offset if the proportion of edges in the prior years graph increases as given by the $\set{E}$ ratio. For example, if a citation graph has a large node count and low proportion of edges in the prior subgraph, the $\lambda$ term will be low. Figure \ref{fig:lambda} illustrates the moderate correlation between proposed GNN model improvement (vs. MLP) and the $\lambda$ parameter.
%A high $\lambda$ parameter score implies a greater benefit from the graph topolgy when using the proposed model. 

\begin{figure}[t]
\begin{minipage}[b]{1.0\linewidth}
  \centering
  \centerline{\includegraphics[width=8.5cm]{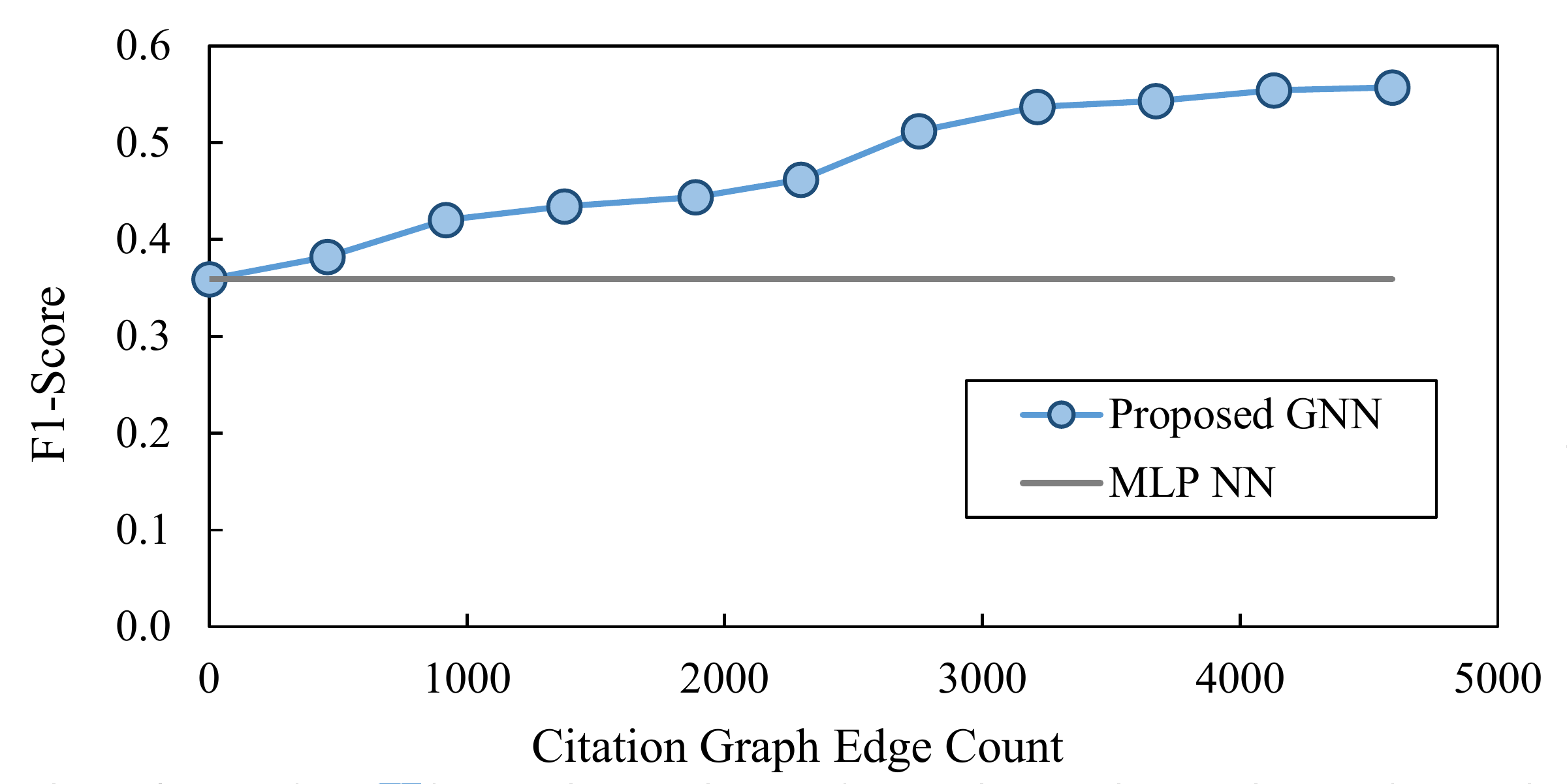}}
%  \vspace{2.0cm}
\end{minipage}
\caption{Results showing the impact of edge connections for predicting the top 10\% of ICML papers. Target year = 2015, 10 year prior graph window.}
\label{fig:edge}
\end{figure}

In another experiment, we start randomly removing edge connections from the graph to see how it affects the predictive behavior. Figure \ref{fig:edge} clearly shows the impact of the graph connections on the proposed GNN model where the F1-score converges to the MLP at zero graph connections. 

\begin{figure}[t]
\begin{minipage}[b]{1.0\linewidth}
  \centering
  \centerline{\includegraphics[width=8.5cm]{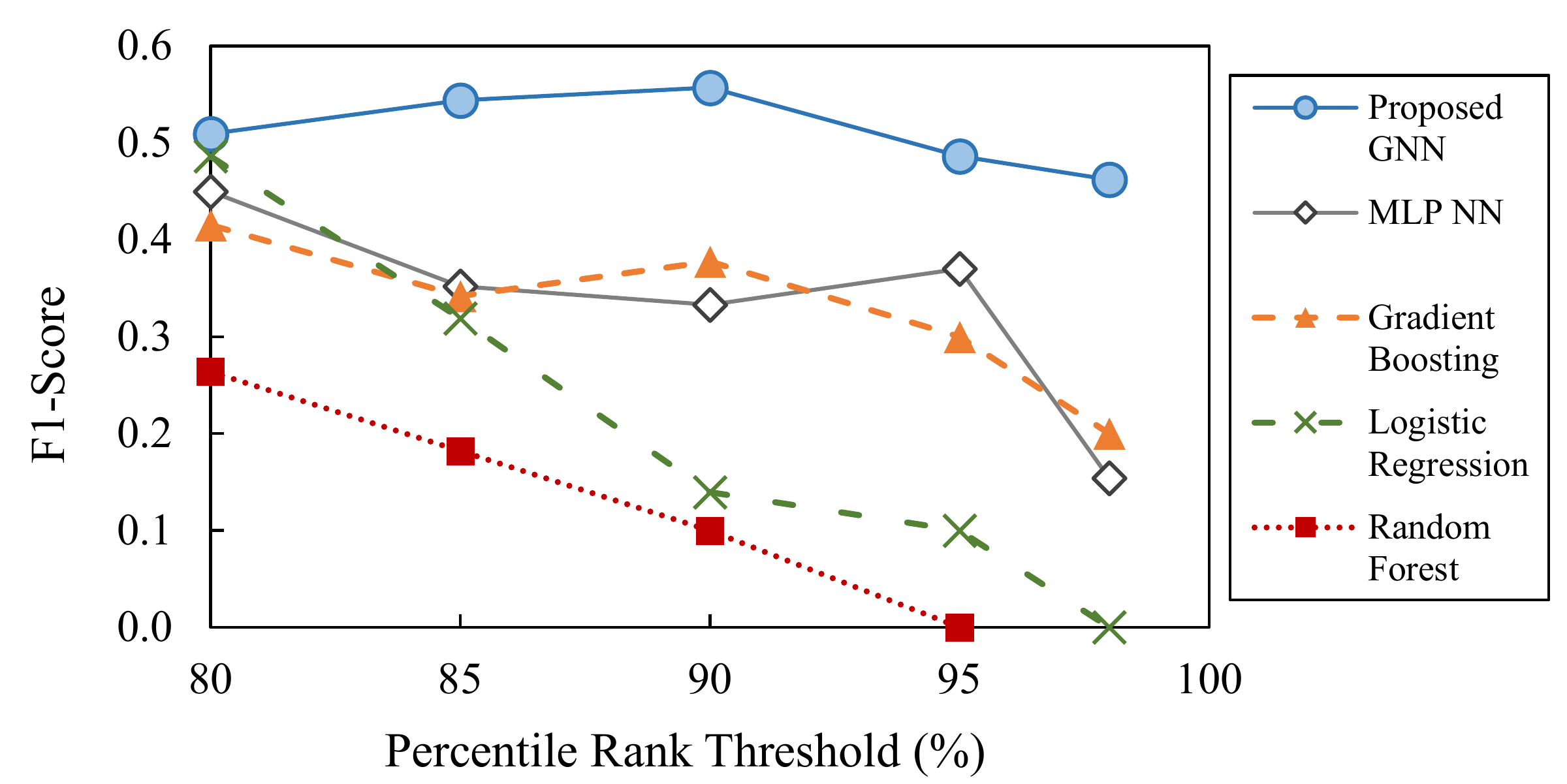}}
%  \vspace{2.0cm}
\end{minipage}
\caption{Comparison of models for predicting ICML papers. Target year = 2015, 10 year prior graph window.}
\label{fig:icml}
\end{figure}

Other classic machine learning models were also evaluated to comparison against the neural network models and were trained on the same sets of input data. Using the ICML 2015 target as an example, Figure \ref{fig:icml} shows that a gradient boosting ensemble method can perform close to the MLP model whereas the logistic regression and random forest models do not predict as well. For this graph, the proposed model shows a consistent F1-score, even up to predicting the top 2\% most cited papers.

\section{Conclusions}
\label{sec:concl}
This paper has demonstrated a structured citation trend prediction approach using a GNN-based architecture that enables trend predictions on key features in recently added nodes for a given citation graph. 
% To exercise the proposed approach, we applied the model across a range of academic citation graphs where the results show a clear advantage over an equivalent MLP in terms of the F1-score. 
It was also shown that graph connectivity matters to how well the GNN approach performs. One area of future work will be investigate if link prediction can be used to enhance node connectivity and increase the predictive power of the model.

%f citation trend prediction due to the recently increased interest in bibliometrics across all research fields. The experiment implemented a set of text, author, and date input features into embedding layers followed by a set GCN layers to achieve an F1-scores better the a parametrically equivalent MLP. 

%\vfill\pagebreak

%section{REFERENCES}
%\label{sec:refs}

\bibliographystyle{IEEEbib}
\bibliography{refs}

\end{document}